\documentclass[manuscript]{acmart}
\AtBeginDocument{%
  }

\copyrightyear{2026}
\acmYear{2026}
\setcopyright{cc}
\setcctype{by}
\acmConference[LAK 2026]{LAK26: 16th International Learning Analytics and Knowledge Conference}{April 27-May 01, 2026}{Bergen, Norway}
\acmBooktitle{LAK26: 16th International Learning Analytics and Knowledge Conference (LAK 2026), April 27-May 01, 2026, Bergen, Norway}
\acmDOI{10.1145/3785022.3785110}
\acmISBN{979-8-4007-2066-6/2026/04}

\usepackage{multirow}
\usepackage{subcaption}

\begin{document}

\title{Modeling Programming Skills with Source Code Embeddings for Context-aware Exercise Recommendation}

\author{Carlos Eduardo P. Silva}
\affiliation{%
  \institution{Departamento de Informática \\ Universidade Federal de Viçosa}
  \city{Viçosa}
  \state{Minas Gerais}
  \country{Brazil}
}
\affiliation{%
  \institution{Instituto Federal de Minas Gerais}
  \city{Ouro Branco}
  \state{Minas Gerais}
  \country{Brazil}
}
\email{carlos.e.silva@ufv.br}

\author{João Pedro M. Sena}
\affiliation{%
 \institution{Instituto de Ciências Matemáticas e de Computação Universidade de São Paulo}
 \city{São Carlos}
 \state{São Paulo}
 \country{Brazil}
 }
 \email{joao.sena@usp.br}

\author{Julio C. S. Reis}
\email{jreis@ufv.br}
\author{André G. Santos}
\email{andresantos@ufv.br}
\affiliation{%
  \institution{Departamento de Informática \\ Universidade Federal de Viçosa}
  \city{Viçosa}
  \state{Minas Gerais}
  \country{Brazil}
}

\author{Lucas N. Ferreira}
\affiliation{%
  \institution{Departamento de Ciência da Computação \\ Universidade Federal de Minas Gerais}
  \city{Belo Horizonte}
  \state{Minas Gerais}
  \country{Brazil}
}
\email{lferreira@dcc.ufmg.br}

\renewcommand{\shortauthors}{Silva, et al.}

\begin{abstract}
In this paper, we propose a context-aware recommender system that models students’ programming skills using embeddings of the source code they submit throughout a course. These embeddings predict students’ skills across multiple programming topics, producing profiles that are matched to the skills required by unseen homework problems. To generate recommendations, we compute the cosine similarity between student profiles and problem skill vectors, ranking exercises according to their alignment with each student’s current abilities. We evaluated our approach using real data from students and exercises in an introductory programming course at our university. First, we assessed the effectiveness of our source code embeddings for predicting skills, comparing them with token-based and graph-based alternatives. Results showed that Jina embeddings outperformed TF-IDF, CodeBERT-cpp, and GraphCodeBERT across most skills. Additionally, we evaluated the system’s ability to recommend exercises aligned with weekly course content by analyzing student submissions collected over seven course offerings. Our approach consistently produced more suitable recommendations than baselines based on correctness or solution time, indicating that predicted programming skills provide a stronger signal for problem recommendation.
\end{abstract}



\maketitle

\section{Introduction}

The growing interest in Computer Science (CS) programs has led universities worldwide to expand the number of seats in related courses, making it increasingly difficult to provide personalized attention to students \cite{highered_2023}. A key challenge arises in introductory programming courses, which serve as foundational prerequisites in the CS curriculum. These courses often enroll large cohorts of students with diverse backgrounds. They also rely on sequential programming assignments that build upon one another, making it especially difficult for students who fall behind to catch up. The lack of personalization in these large heterogeneous classes might lead to high dropout rates, which are particularly harmful for public universities funded by the government.  

Personalized programming assignments can help address these issues by accommodating students with varying levels of expertise, allowing them to engage with the course more effectively \cite{medeiros_2019, becker_2019}. However, designing enough programming problems and assigning them individually remains a significant challenge for instructors. Recommender systems offer a promising solution, modeling individual students’ programming expertise and suggesting customized assignments accordingly \cite{pereira_2021, rahman_2022}. Several programming exercise recommender systems have been proposed in the literature \cite{zaffalon_2022, vesin_2022, pereira_2023}, with recent ones employing content-based approaches based on neural network embeddings \cite{muepu_2025} to represent code submitted by students. Typically, these systems do not explicitly model learners’ knowledge during a course; instead, they recommend problems based on simple similarity between problems not yet seen by students and those they have already solved. 

In this paper, we propose a context-aware recommendation system \cite{bakhshizadeh_2024} that directly models a student’s learning context, defined as their programming skill, through the sequence of source code submissions made during the course. Our objective is to generate personalized programming assignments within a formal academic course setting, improving personalized education and potentially reducing dropout. Each student submission in our system is encoded using embeddings generated by a BERT-based source code model, Jina \cite{sturua_2024}, which supports 30 programming languages. To capture the student’s current learning context, we select the embedding closest to the centroid of all submissions in the most recent lab. This vector provides a dynamic representation of the student’s programming ability. The context vector is then used as input to multiple multiclass classifiers, each trained to predict the student’s proficiency in a specific programming topic (e.g., mathematics, conditionals, repetition, arrays). The resulting set of predicted skills forms a profile vector, which we compare via cosine similarity to embeddings of unseen programming exercises labeled with the same skill categories. By ranking exercises according to similarity, we generate personalized programming assignments, presenting the top-$k$ exercises in descending order of match quality.

To train the skill classifiers and evaluate our method, we curated a dataset of source code submissions from the last seven offerings of the introductory programming course (INF110) in the CS curriculum of the Universidade Federal de Viçosa, a public research university in the state of Minas Gerais, Brazil. This dataset includes 12,912 C++ source code submissions from 253 students, covering 112 programming problems ($\approx$ 53 per semester). The dataset is fully anonymized and contains no personally identifying information.

We conducted two experiments to evaluate our system. First, we measured the effectiveness of Jina embeddings for skill prediction compared to other approaches: TF-IDF, CodeBERT-cpp \cite{zhou_2023}, and GraphCodeBERT \cite{guo_2021}. Results showed that Jina substantially outperforms TF-IDF and slightly outperforms the other BERT-based methods. While Jina’s performance improvement over other BERT-based embeddings is modest, it is important to note that Jina natively supports 30 programming languages, whereas GraphCodeBERT supports 7 and CodeBERT-cpp supports only C++.

Our second experiment evaluated the effectiveness of our skill-based recommendation system in suggesting suitable problems. A problem is considered suitable if it requires only the skills introduced between the student’s last submission and the current class. We compared our skills-based ranking against two alternative signals commonly available from online judges: solution time and solution correctness. For each signal, we also tested two strategies for representing students’ submissions: (1) the average of their embeddings and (2) the embedding of the submission closest to the centroid. The second strategy is motivated by the observation that averaging embeddings may produce vectors that do not correspond to valid exercises, while selecting the closest embedding ensures that students are represented by an actual code solution. Results showed that recommendations based on skills outperform those based on the alternative signals. Moreover, representing students by the embedding closest to the centroid of their most recent lab submissions outperforms all other representation strategies.

\section{Related Work} \label{sec:relatedwork}

Our work primarily concerns Educational Recommender Systems and, consequently, Learner Modeling. This section briefly reviews previous studies in these areas related to programming exercise recommendation.

\subsection{Programming Learning Modelling}

Learner Modelling techniques are a key component of Educational Recommender Systems, aiming to approximate a student's knowledge state, skills, or needs based on their past interactions with the educational system \cite{ain_2024}. Traditional learner modeling approaches in the computer programming domain include the Elo Rating System (ERS), Bayesian Knowledge Tracing (BKT), and Item Response Theory (IRT) \cite{vesin_2022}. For example, Fori{\v{s}}ek \cite{forisek2009using} introduced an IRT-based rating system tailored for programming competitions, modeling both contestants’ abilities and task properties (difficulty and discrimination) within a unified probabilistic framework. Mangaroska et al. \cite{mangaroska_2019} presented an adaptive assessment framework for programming courses based on a modified Elo rating algorithm, in which learners and coding exercises are co-rated using non-binary performance signals such as attempts and success ratios.

One major drawback of these traditional learner modeling techniques is their heavy reliance on manual feature engineering and hyperparameter tuning \cite{pelánek_2017, liu_2022}. Consequently, recent approaches have explored deep learning methods to learn vector representations (embeddings), aiming to automatically capture complex semantic features from both problem statements and students’ solutions. For example, Azcona et al. \cite{azcona_2019} proposed a model called \emph{user2code2vec} to build profiles of programming students using embeddings of Python code submissions throughout a course, extracted by a simple MLP network. Similarly, Yoder et al. \cite{yoder_2022} adapted the \emph{Code2Vec} model, which uses paths in an Abstract Syntax Tree (AST) to create vector representations, for predicting final grades in a Java programming course.

Our recommender system also uses embeddings as a learner modeling approach. However, the main difference from previous work is that we use Jina \cite{sturua_2024}, a multilingual transformer network, to extract embeddings from source code submitted by students, and we use these embeddings to predict students' programming skills.

\subsection{Programming Exercise Recommender Systems}

Several systems have been proposed for programming exercise recommendation, with varying educational goals, learner models, and recommendation techniques \cite{zaffalon_2022, vesin_2022, pereira_2023}. For example, Vesin et al. \cite{vesin_2022} proposed a recommendation framework for programming courses based on a modified Elo-rating algorithm, in which students and exercises are co-rated using performance signals such as correctness, number of attempts, and time to solution. To recommend an exercise, the system selects problems whose estimated difficulty is close to the student’s current rating. Muepu et al. \cite{muepu_2025} proposed a purely content-based recommender that combines learned embeddings and AST features to represent problems, enabling recommendations based on similarity to previous problems solved by the students.

Our recommender system is similar to previous content-based systems, as we also recommend problems based on the similarity of learned source code embeddings. However, it differs from these earlier methods because we do not simply recommend problems similar to previously solved problems; instead, we compute a dynamic learning context for the student and recommend problems similar to this context.

\section{Methodology} \label{sec:methodology}

We propose a context-aware recommendation system with the goal of producing personalized homework programming assignments for students. For that, we assume a course structure organized weekly, where each week consists of (1) a lecture, (2) a lab session, and (3) a homework assignment. During the lecture, students are introduced to programming topics, and in the labs, they are assigned a set of introductory programming problems to practice the discussed techniques. In the homework assignments, they have a new set of programming problems not seen in the lab, some with a similar difficulty and some slightly harder, to support both students who need more practice and those who need more challenge. Table \ref{tab:course_structure} illustrates three weeks of a course that follows our structure, where each problem (P1, P2, P51, etc.) is annotated with difficulty ratings 0 (topic not covered), 1 (easy), 2 (medium), and 3 (hard) for different programming topics. 

\begin{table*}[!htbp]
 \
\caption{Weekly structure of a programming course including Lectures, Labs, and Homework tasks. Each problem is annotated with difficulty ratings from 0 to 3 for different programming topics.}
\label{tab:course_structure}
\begin{tabular}{@{}lllcccccccc@{}}
\toprule
\multicolumn{3}{c}{} & \multicolumn{8}{c}{\textbf{Programming Topics}} \\
\cmidrule(lr){4-11}
\multicolumn{3}{c}{} & \textbf{Math} & \textbf{Cond.} & \textbf{Rep.} & \textbf{Array} & \textbf{Matrix} & \textbf{Func.} & \textbf{String} & \textbf{Struct} \\
\midrule

\multirow{5}{*}{\textbf{W2}} & \textbf{Lec} &  & X & - & - & - & - & - & - & - \\ \cline{2-11}
 & \multirow{2}{*}{\textbf{Lab}} & P1 & 1 & 0 & 0 & 0 & 0 & 0 & 0 & 0 \\
 &  & P2 & 2 & 0 & 0 & 0 & 0 & 0 & 0 & 0 \\ \cline{2-11}
 & \multirow{2}{*}{\textbf{Hw}} & P51 & 2 & 0 & 0 & 0 & 0 & 0 & 0 & 0 \\
 &  & P52 & 3 & 0 & 0 & 0 & 0 & 0 & 0 & 0 \\
\midrule

\multirow{6}{*}{\textbf{W3}} & \textbf{Lec} &  & - & X & - & - & - & - & - & - \\ \cline{2-11}
 & \multirow{3}{*}{\textbf{Lab}} & P3 & 1 & 1 & 0 & 0 & 0 & 0 & 0 & 0 \\
 &  & P4 & 2 & 2 & 0 & 0 & 0 & 0 & 0 & 0 \\
 &  & P5 & 3 & 1 & 0 & 0 & 0 & 0 & 0 & 0 \\ \cline{2-11}
 & \multirow{2}{*}{\textbf{Hw}} & P53 & 2 & 2 & 0 & 0 & 0 & 0 & 0 & 0 \\
 &  & P54 & 3 & 3 & 0 & 0 & 0 & 0 & 0 & 0 \\
\midrule

\multirow{7}{*}{\textbf{W4}} & \textbf{Lec} &  & - & - & X & - & - & - & - & - \\ \cline{2-11}
 & \multirow{4}{*}{\textbf{Lab}} & P6 & 1 & 1 & 1 & 0 & 0 & 0 & 0 & 0 \\
 &  & P7 & 1 & 2 & 2 & 0 & 0 & 0 & 0 & 0 \\
 &  & P8 & 3 & 3 & 1 & 0 & 0 & 0 & 0 & 0 \\
 &  & P9 & 2 & 3 & 3 & 0 & 0 & 0 & 0 & 0 \\ \cline{2-11}
 & \multirow{2}{*}{\textbf{Hw}} & P57 & 2 & 3 & 2 & 0 & 0 & 0 & 0 & 0 \\
 &  & P58 & 3 & 2 & 3 & 0 & 0 & 0 & 0 & 0 \\
\bottomrule \\
\end{tabular}
\end{table*}

In this hypothetical course, students are exposed to Math, Conditionals, and Repetition in lectures of weeks 2, 3, and 4, respectively. The lab in week 2 presents two problems, P1 and P2, requiring only Math with difficulty levels of 1 and 2. In week 3, the lab includes three problems, P3, P4, and P5, combining Math and Conditionals. Week 4 incorporates all three topics with varying levels of difficulty. After each week’s lab, students are assigned two homework problems: one at the same difficulty level as those seen in the lab and one that is harder. Homework assignments are typically the same for all students, regardless of their progress in the lab. This fixed assignment becomes problematic as students diverge in understanding throughout the course, since their mastery of topics differs substantially.

\subsection{Context-aware Student Modeling}

Our goal in this paper is to model students’ progress during lab sessions to recommend personalized homework problems. Our approach consists of modeling a student $e$’s programming skills, which we also refer to as their learning context, by the following sequence:

\begin{equation*}
S^{\{e\}} = \{s_{11}^{\{e\}[1]}, s_{12}^{\{e\}[1]}, s_{12}^{\{e\}[2]}, ..., s_{21}^{\{e\}[1]}, ...,  s_{m1}^{\{e\}[1]},  ..., s_{mn}^{\{e\}[T]}\},    
\end{equation*}

where $s_{ij}^{\{e\}[k]}$ represents the $k$-th submission of student $e$ to problem $j$ of lab $i$, $m$ is the total number of lab sessions in the course, $n$ is the number of problems in the last lab, and $T$ is the number of submissions made to the last problem of the last lab. We explicitly consider multiple submissions to the same problem (see $s_{12}^{\{e\}[1]}, s_{12}^{\{e\}[2]}$ above), since it is common for students to require several attempts before solving correctly. Thus, this sequence differs for students who solved the same set of problems in a different order or with a different number of attempts.

Since each element $s_{ij}^{\{e\}[k]}$ is source code, we convert it into a feature vector using a BERT-based embedding model. In particular, we use \emph{jina-embeddings-v2-base-code}\footnote{\url{https://huggingface.co/jinaai/jina-embeddings-v2-base-code}}, a state-of-the-art model that supports 30 programming languages and a context size of 8192 tokens. This context size is sufficient to represent student programs in general. Additionally, the model supports Matryoshka Representation Learning \cite{kusupati_2022}, which allows reducing the embedding dimensionality without compromising performance. By representing students’ solutions with \emph{jina-embeddings-v2-base-code}, our recommender system is effectively programming-language-agnostic, supporting the same 30 languages.

\subsection{Recommending Programming Problems}

We formulate our recommendation problem as predicting the programming skills students currently have from their submission embeddings and selecting the $k$ homework problems most similar in terms of required skills. We considered the eight programming skills shown in Table \ref{tab:course_structure}, with categorical labels from 0 (topic not covered), 1 (easy), 2 (medium), and 3 (hard). Formally, our problem consists of finding a function $\hat{y}^{\{e\}} = f(S^{\{e\}})$ that predicts the skills $\hat{y}^{\{e\}} \in \{0,1,2,3\}$ of a student $e$ given their submission sequence $S^{\{e\}}$. 

Since we are in a low-data regime, we do not consider sequential models such as Recurrent Neural Networks or Transformers as $f$. Instead, we summarize the sequence $S^{\{e\}}$ into a single context vector $x^{\{e\}}$ by selecting the embedding of the submission closest to the centroid of the last lab attended by the student. This assumes that using a real submission is more representative than averaging embeddings, which might yield a vector that does not correspond to a valid exercise, and that recent code better reflects the student’s current skills. Our skill prediction model is therefore defined as a classification Multilayer Perceptron (MLP):

\[
\hat{y}^{\{e\}} = f(x^{\{e\}}) = MLP_{\theta}(x^{\{e\}}),
\]

where $\theta$ is the set of model parameters learned from data. Since we have eight programming skills, each representing a separate multiclass problem (labels $\{0,1,2,3\}$), we train eight independent MLPs, one per skill.

To recommend homework problems, we retrain our eight $MLP_{\theta}$ models after each lab session $l$ and use them to predict the skills of homework problems $p_c \in C$, where $C$ is the set of recommendation problems not used in any lab. Each candidate problem $p_c$ is represented by the Jina embedding of a reference source code provided by the instructors. The final recommendation list is produced via cosine similarity between the student’s predicted skill vector and the predicted skill vectors of each homework problem $p_c \in C$. Problems are presented in decreasing order of similarity, assuming that higher similarity corresponds to easier and more suitable problems for a given student.

The predicted skills are the only signal we use to rank programming problems. For simplicity, we do not consider additional learning signals such as lecture attendance or extra practice submissions. It is important to note that the auto-grader is used only for automation; our methodology does not rely on it.

\subsection{Dataset}
\label{subsec:dataset}

To train our MLP models, we collected a dataset of $12{,}912$ student submissions to problems from the introductory programming course (INF110) in the CS curriculum of the Universidade Federal de Viçosa, a public research university in the state of Minas Gerais, Brazil. This course is offered once per year and follows a semester calendar with a weekly structure similar to that shown in Table~\ref{tab:course_structure}. The dataset contains submissions to $112$ different programming problems from the last seven editions of the course: 2018, 2019, 2021, 2022, 2023, 2024, and 2025. The 2020 edition was excluded due to disruptions caused by the COVID-19 pandemic.

The problems cover eight introductory programming topics: Math, Conditionals, Repetition, Arrays, Matrices, Functions, Strings, and Structures. Each submission is a tuple $(e, i, p, s)$, where $e$ is an anonymous student ID, $i$ is the lab index, $p$ is the problem index, and $s$ is the source code. All submissions are in C++, since our course introduces students to the fundamentals of procedural programming in that language. Moreover, students’ solutions are graded by an auto-grader using hand-crafted test cases to determine correctness.

All problems were specifically designed for this course by a team of three professors who typically teach it. In general, problems follow a structure commonly found in programming contest settings, such as the International Collegiate Programming Contest (ICPC) \cite{trotman_2008}. Each problem begins with a title, followed by a descriptive statement, and concludes with input/output specifications and constraints. On average, problem statements are 238.71 words long ($\pm$ 108.47), and code solutions contain 140.08 ($\pm$ 75.01) tokens.

To evaluate our context representation based on source code embeddings, we systematically labeled each problem in the dataset with difficulty levels (0--3) across the eight programming topics in Table~\ref{tab:course_structure}. Labeling was performed consistently by a professor who has taught the course for more than 10 years, including all seven years of data collection. To illustrate the approach, the labeling criteria for the Conditionals topic are as follows: $1$ if the solution requires a simple \textit{if} statement, $2$ if it requires a basic \textit{if--else} statement, and $3$ if it requires a nested \textit{if--else} structure.

Finally, the data collection process and the experiments presented in this paper were approved by the Ethics Committee of the Universidade Federal de Viçosa under project number CAAE 75327023.0.0000.5153. 

\section{Experimental Results} \label{sec:results}

\begin{table*}[!htbp]
    \centering
    \caption{Average accuracies of MLP skill classifiers trained with different contextual representation methods across 11 lab sessions.}
    \label{tab:experiment_1}
    \begin{tabular}{lcccc}
        \toprule
        & Jina & TF-IDF & CodeBERT-cpp & GraphCodeBERT \\
        \midrule
        Math & \textbf{0.75 $\pm$ 0.29} & $0.66 \pm{0.28}$ & $0.68 \pm{0.36}$ & $0.61 \pm{0.35}$ \\
        Conditional & \textbf{0.64 $\pm$ 0.18} & $0.52 \pm{0.17}$ & $0.63 \pm{0.14}$ & $0.61 \pm{0.16}$ \\
        Repetition & $0.37 \pm{0.13}$ & $0.32 \pm{0.11}$ & \textbf{0.41 $\pm$ 0.17} & $0.39 \pm{0.14}$ \\
        Array & \textbf{0.52 $\pm$ 0.20} & $0.45 \pm{0.13}$ & \textbf{0.52 $\pm$ 0.22} & $0.47 \pm{0.19}$ \\
        Matrix & \textbf{0.53 $\pm$ 0.37} & $0.46 \pm{0.34}$ & $0.51 \pm{0.35}$ & $0.51 \pm{0.33}$ \\
        Function & \textbf{0.80 $\pm$ 0.26} & $0.54 \pm{0.37}$ & $0.75 \pm{0.28}$ & $0.74 \pm{0.29}$ \\
        String & $0.89 \pm{0.23}$ & $0.86 \pm{0.19}$ & \textbf{0.90 $\pm$ 0.22} & $0.89 \pm{0.23}$ \\
        Struct & \textbf{0.98 $\pm$ 0.04} & $0.87 \pm{0.25}$ & $0.90 \pm{0.26}$ & $0.89 \pm{0.24}$ \\ \midrule
        Mean & \textbf{0.69 $\pm$ 0.19} & $0.59 \pm{0.19}$ & $0.66 \pm{0.17}$ & $0.64 \pm{0.18}$ \\
        \bottomrule
    \end{tabular}
\end{table*}

We evaluate our context-aware recommendation system through two experiments. First, we assess the effectiveness of our student context representation \( x^{\{e\}} \) by measuring skill prediction accuracy using a multilayer perceptron (MLP) model for each skill. Second, we evaluate our skill-based recommender against other recommendation signals, namely solution time and solution correctness. These are alternative labels that can also be used to rank homework problems and are relatively easy to retrieve from online judges. As part of this experiment, we also consider multiple strategies to summarize the student learning context from their code embeddings. 

\subsection{Student Context Representation}

In the first experiment, we evaluate the effectiveness of Jina embeddings as predictors of student skills, comparing them with embeddings from other token-based methods, TF-IDF and CodeBERT-cpp, as well as GraphCodeBERT, a graph-based approach. For all four methods, we trained eight MLP multiclass classifiers with a single hidden layer of 100 neurons, using gradient descent with consistent hyperparameters: 200 epochs, a learning rate of \(\eta = 0.001\), and L2 regularization with \(\lambda = 0.1\). The embeddings from all BERT-based methods (Jina, CodeBERT-cpp, and GraphCodeBERT) were represented as 768-dimensional vectors. Additionally, Jina incorporates five LoRA (Low-Rank Adaptation) adapters \cite{hu_2021} for specific tasks: retrieval-query, retrieval-passage, separation, classification, and text-matching. Given the nature of our recommendation task, we used the text-matching adapter. As TF-IDF embeddings have a dimensionality equal to the vocabulary size—i.e., the number of unique words across all documents—we reduced this dimensionality by selecting only the most frequent terms, matching the resulting vectors to the 768-dimensional size of the BERT-based embeddings. 

To simulate how the system would be used in practice, predicting skills for unseen students and problems in the future, we adopted a time-sensitive leave-one-out approach for dataset splitting. Specifically, we trained our models on the correct submissions from all lab sessions from 2018 to 2024, and evaluated them on the 11 lab sessions of 2025. Since each of the eight skills represents an individual multiclass problem, to improve generalization we enforced a balanced dataset when training each model by downsampling to the size of the minority class. Table \ref{tab:experiment_1} reports the average accuracies of each skill for each context representation method across the 11 lab sessions of 2025.

Table \ref{tab:experiment_1} shows that Jina embeddings consistently outperform the other contextual representation methods, achieving the highest mean accuracy (0.69) and leading in six of the eight skills, including Math (0.75), Conditional (0.64), Function (0.80), and Struct (0.98). CodeBERT-cpp provides competitive results, surpassing Jina on Repetition (0.41 vs. 0.37) and slightly on String (0.90 vs. 0.89), while tying with Jina on Array (0.52). GraphCodeBERT does not improve over CodeBERT-cpp, and TF-IDF lags behind all neural methods, confirming the importance of contextual embeddings over frequency-based representations. Overall, Jina not only achieves the best performance across most skills but also supports a broader range of programming languages (30), compared to the more limited coverage of CodeBERT-cpp (C++ only) and GraphCodeBERT (seven languages).

\subsection{Ranking Metrics}

In our second experiment, we evaluate predicted programming skills as a ranking metric for recommending programming problems. Since our recommendation problem has a temporal component, the solution time MLP model is retrained after each lab session. The goal of this experiment is to assess how our recommendation algorithm, guided by skill predictions from source code embeddings, learns to suggest suitable homework problems for students, where suitable is defined as problems that only require skills in the interval between the student’s last submission and the current class. To this end, we compare our categorical skill predictions against two other ranking metrics: solution time and solution correctness. The solution time is the time in seconds a student takes to solve a problem, which can normally be recorded by online judges and used as a proxy for problem difficulty. Solution correctness, on the other hand, is a binary label indicating whether a given submission is correct—this is the main feature tracked by most online judges.

\begin{figure*}[!htbp]
    \centering
    \begin{subfigure}{0.39\textwidth}
        \centering
        \includegraphics[width=\linewidth]{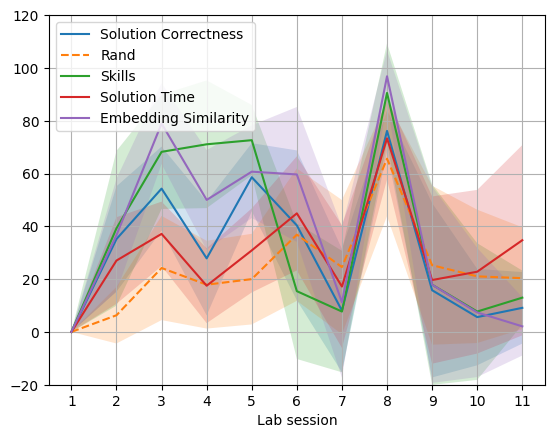}
        \Description{Line chart showing adequate exercises for all methods in all class, using all submissions and average context.}
        \caption{Average of all submissions.}
    \end{subfigure}
    \begin{subfigure}{0.39\textwidth}
        \centering
        \includegraphics[width=\linewidth]{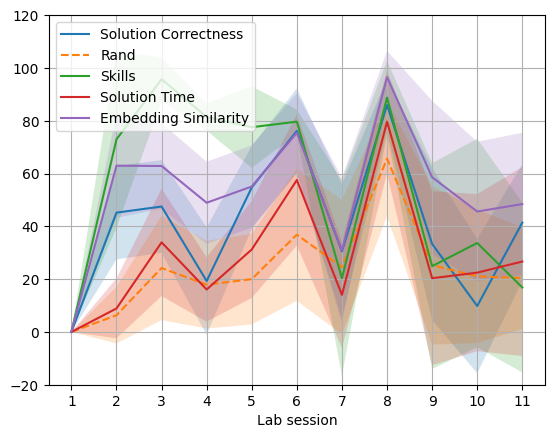}
        \Description{Line chart showing adequate exercises for all methods in all class, using last lab submissions only and average context.}
        \caption{Average of last lab submissions.}
    \end{subfigure}
    \begin{subfigure}{0.395\textwidth}
        \centering
        \includegraphics[width=\linewidth]{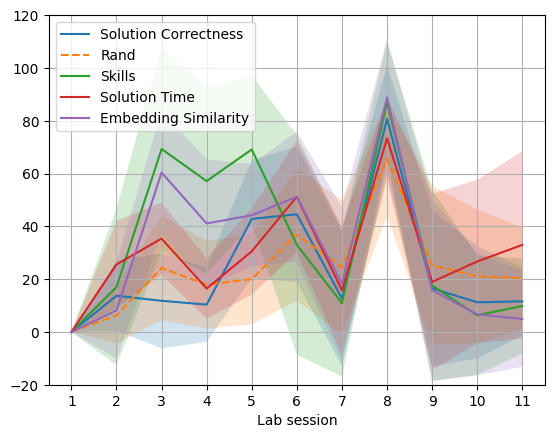}
        \Description{Line chart showing adequate exercises for all methods in all class, using all submissions and centroid context.}
        \caption{Solution closest to the centroid of all submissions.}
    \end{subfigure}
    \begin{subfigure}{0.395\textwidth}
        \centering
        \includegraphics[width=\linewidth]{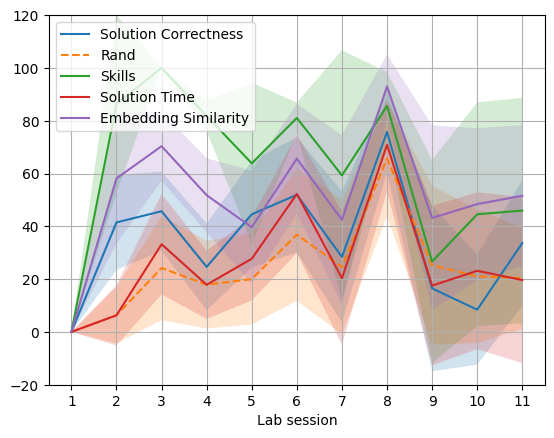}
        \Description{Line chart showing adequate exercises for all methods in all class, using last lab submissions only and centroid context.}
        \caption{Solution closest to the centroid of last lab submissions.}
    \end{subfigure}
    
    \caption{Average percentage of suitable exercises (the higher, the better) of different ranking metrics (Skills, Solution Time, and Solution Correctness) with varying strategies for summarizing the submissions into a single embedding vector.}
    \label{fig:ranking_metrics}
\end{figure*}

As part of this experiment, we also evaluate two strategies for summarizing student submissions into a single embedding vector: (1) the average of the student’s submission embeddings, and (2) the embedding of the solution closest to the student centroid, as represented by Jina embeddings. Since our recommendation problem has a temporal component, for both strategies, we compare summarizing across all submissions versus only submissions from the most recent lab.

Figure \ref{fig:ranking_metrics} shows the percentage of suitable recommended exercises produced by our system when guided by the three different ranking metrics: solution time, solution correctness, and skills. In this figure, the higher the percentage, the better. Each ranking metric was evaluated under the four summarization strategies. For comparison, we also report the percentage of suitable recommended exercises a theoretical random method would achieve, in order to evaluate how much the recommendation systems are actually learning from the data. For the variations with solution time and solution correctness, we trained MLP predictors with the same size and configuration as those used for skill prediction.

The results show that solution time and correctness as ranking metrics are not strongly affected by the summarization strategy. This is likely due to the low variance of solution time and correctness across labs. Since each lab introduces new content, problems become more complex in terms of the number of skills required, but the difficulty of each individual skill remains relatively similar across labs. Moreover, students have the same allotted lab time (1 hour and 40 minutes) in all sessions, which limits the variance of solution time and correctness throughout the course. Our skill-based metric, on the other hand, consistently performs better when using only the solution closest to the last lab centroid, especially in the second half of the course. These results suggest that predicted skills are a stronger recommendation signal for programming problems, as they better capture differences in students’ learning contexts, particularly as the number of required programming skills increases. Overall, our skill-based ranking metric—when combined with a student context based solely on the solution closest to the last lab centroid—outperformed all other methods.

\section{Conclusion and Future Work} \label{sec:conclusion}

This paper introduced a context-aware recommender system for programming assignments designed for formal academic courses. By modeling students’ skills from their code submissions and using embeddings for skill prediction, the system recommends exercises that align with each student’s learning trajectory. Experiments on a new multi-year dataset show that skill-based recommendations outperform those based on correctness or completion time, and that Jina embeddings provide consistent improvements while supporting a broader range of programming languages.

The contributions of this paper represent an initial step toward personalized programming assignments via recommender systems based on source code embeddings. As future work, we plan to conduct user studies to assess the perceived adequacy and difficulty of recommendations and to measure their impact on student learning outcomes. These studies will be conducted with a new cohort of students arriving next year. We also intend to extend the approach to other courses and disciplines, exploring the generalizability of embedding-based skill modeling for personalized education.

\begin{acks}
This work was partially supported by grants from INCT-TILDIAR, CAPES, CNPq, FAPEMIG, and Instituto Federal de Educação, Ciência e Tecnologia de Minas Gerais.
\end{acks}

\bibliographystyle{ACM-Reference-Format}
\bibliography{references}
\end{document}